\documentclass{article}
\PassOptionsToPackage{numbers, compress}{natbib}
\usepackage[final]{neurips_2021}

\usepackage{natbib}

\usepackage[utf8]{inputenc} % allow utf-8 input
\usepackage[T1]{fontenc}    % use 8-bit T1 fonts
\usepackage{hyperref}       % hyperlinks
\usepackage{url}            % simple URL typesetting
\usepackage{booktabs}       % professional-quality tables
\usepackage{amsfonts}       % blackboard math symbols
\usepackage{nicefrac}       % compact symbols for 1/2, etc.
\usepackage{microtype}      % microtypography
\usepackage{xcolor}         % colors

\title{On Combining Expert Demonstrations in Imitation Learning via Optimal Transport}
\author{%
  Ilana Sebag \\
  Department of Computer Science\\
  University College London, UK
   \And
   Samuel Cohen \\
   Centre for Artificial Intelligence\\
   University College London, UK 
   \AND
   Marc Peter Deisenroth \\
   Centre for Artificial Intelligence\\
   University College London, UK 
}

\usepackage{color, colortbl}
\newcommand{\norm}[1]{\left\lVert#1\right\rVert}
 
\usepackage{booktabs}               % professional-quality tables

\usepackage{caption}
\usepackage{subcaption}
\usepackage{amsmath,amsfonts,bm}
\usepackage{multirow}
\usepackage{tabularx}
\usepackage{graphicx}
\usepackage{adjustbox}
\usepackage{amsthm}
\usepackage{xspace}

\newcommand{\mset}[1]{\left\{\kern-.5em\left\{ #1 \right\}\kern-.5em\right\}}
\newcommand{\mmset}[1]{\{\kern-.4em\{ #1 \}\kern-.4em\}}

\newcommand{\abs}[1]{\left\vert#1\right\vert}

\def\1{\bm{1}}

\def\vtheta{{\bm{\theta}}}

\def\va{{\bm{a}}}

\def\vs{{\bm{s}}}

\def\vx{{\bm{x}}}

\def\vec1{{\bm{1}}}

\DeclareMathAlphabet{\mathsfit}{\encodingdefault}{\sfdefault}{m}{sl}
\SetMathAlphabet{\mathsfit}{bold}{\encodingdefault}{\sfdefault}{bx}{n}

\def\gA{{\mathcal{A}}}

\def\gM{{\mathcal{M}}}

\def\gS{{\mathcal{S}}}

\newcommand{\R}{\mathbb{R}}

\usepackage[vlined]{algorithm2e}

\usepackage{amsmath}
\usepackage{amsmath}
\usepackage{amssymb}
\usepackage{mathtools}

\usepackage{amssymb,graphicx,color}
\usepackage{amsfonts}
\usepackage{latexsym}
\usepackage{amsmath}

\usepackage{booktabs}               % professional-quality tables
\usepackage{amsfonts}               % blackboard math symbols
\usepackage{nicefrac}               % compact symbols for 1/2, etc.
\usepackage{microtype}              % microtypography
\usepackage{soul}

\usepackage{geometry}
\usepackage{listings}
\usepackage{graphicx}
\usepackage{enumitem}

\usepackage{amsmath}
\usepackage{amssymb}
\usepackage[frozencache,cachedir=.]{minted}

\usepackage{textcomp}
\begin{document}

\maketitle
%%%%%%%%%%%%%%%
\begin{abstract}

% 1. what is the problem we consider?
% 2. why is it relevant? 
% 3. how do people normally solve this?
% 4. what are the shortcomings of these methods?
% 5. how do we address these shortcomings (key idea only)
% 6. what do we get out of it as a result?

Imitation learning (IL) seeks to teach agents specific tasks through expert demonstrations. One of the  key approaches to IL is to define a distance between agent and expert and to find an agent policy that minimizes that distance. Optimal transport methods have been widely used in imitation learning as they provide ways to measure meaningful distances between agent and expert trajectories. However, the problem of how to optimally combine multiple expert demonstrations has not been widely studied. The standard method is to simply concatenate state (-action) trajectories, which is problematic when trajectories are multi-modal. We propose an alternative method that uses a multi-marginal optimal transport distance and enables the combination of multiple and diverse state-trajectories in the OT sense, providing a more sensible geometric average of the demonstrations. Our approach enables an agent to learn from several experts, and its efficiency is analyzed on OpenAI Gym control environments and demonstrates that the standard method is not always optimal.

\end{abstract}

\section{Introduction}

Imitation learning (IL) techniques aim to mimic expert behaviour in a given task: an expert provides us with a set of demonstrations and the agent uses them to recover the expert's policy. Over the last few years,  IL has been gaining more attention due to a combination of advances in reinforcement learning and deep learning. IL techniques have been used in robotics applications \cite{stepputtis2019imitation, kim2021transformerbased, karnan2021voila} and natural language processing \cite{wu2021textgail, zheng2019simultaneous}. Typical approaches can be divided into two main categories: behavioural cloning (BC) \cite{6796843, goecks2020integrating, 10.5555/2976456.2976601, florence2021implicit} and inverse reinforcement learning (IRL) \cite{Englert2013,bighashdel2021deep, jarboui2021offline, jarboui2021generalised, Ng00algorithmsfor, 10.1145/279943.279964}. The goal in both cases is to find a learned behaviour that matches demonstrated behaviour. In BC, we require state and action observations and then find a (supervised) policy that maps from states to corresponding actions. In IRL, the requirement for observed actions is relaxed, and it is possible to find learned policies that generate state trajectories that are similar to state trajectories observed in demonstrations. Here, we need mechanisms to compare trajectories, and optimal transport (OT) provides us with the mathematical tools to do this in a principled way. 

OT has had contributions in the imitation learning field \cite{xiao2019wasserstein, papagiannis2020imitation} as it allows to compute a discrepancy between discrete probability measures, which can for instance be an expert demonstration and an agent's trajectory rollout. However, the use of multiple expert demonstrations and the optimal way to combine them within a IRL algorithm is still under-explored. The most common way is to concatenate the multiple expert state-action trajectories and subsample a state(-action) trajectory that is used as the expert demonstration \cite{DBLP:journals/corr/abs-2006-04678, zhao2020compressed}. This approach might be sub-optimal when the expert demonstrations and the corresponding trajectories are diverse as the diversity of these trajectories in a single dataset will be considered as noise. 

In this work, we study an alternative way to deal with multiple expert demonstrations by using multi-marginal optimal transport tools. We build two models based on the primal Wasserstein imitation learning (PWIL) algorithm presented in \citet{DBLP:journals/corr/abs-2006-04678}. This work introduces an IRL method leveraging pseudo-rewards computed using a greedy optimal transport distance. Our approach is built on PWIL. However, we consider a different choice of metrics between trajectories, namely sliced OT distances \cite{kolouri2018sliced, cohen2021sliced}, and we consider different methods for combining demonstrations. 

We provide empirical results for both methods on two OpenAI Gym control environments.

\section{Background}

We now introduce the background necessary to define imitation rewards via optimal transport. 

\subsection{Sliced Optimal Transport}
\label{sot}

Optimal transport tools allow us to define distances between probability measures. In this work, we consider discrete measures of the form $\mu = \sum_{i=1}^T \delta_{\vx_t} \in \gM(\R^d)$. Sliced optimal transport distances were proposed in order to reduce the limiting computational complexity of standard OT distances.

Given two discrete probability measures $\mu, \nu \in  \gM(\R^d)$, one can define a distance between them by averaging 2-Wasserstein distances projected onto various axis. The (squared) sliced Wasserstein distance is defined as 
\begin{align}
\mathcal{SW}_2^2(\mu, \nu) = \int_{S^{d-1}} \mathcal{W}^2_2 (  \mathcal{P}_{\vtheta_k \# \mu}, \mathcal{P}_{\vtheta_k\# \nu}) d\vtheta,
\label{sw}
\end{align}
where $S^{d -1} = \{\vtheta \in \mathbb{R}^{d} :  \norm{\vtheta} =1\}$ is the $d$-dimensional sphere in $ \mathbb{R}^{d}$, $\mathcal{P}_{\vtheta}(\vx) = \langle \vtheta,\vx\rangle$ is the linear projection operator, $d\vtheta$ is the normalized uniform measure on the sphere satisfying $\int_{S^{d-1}} d\vtheta =1$ and $\mathcal{W}_2^2$ is the squared 2-Wasserstein distance on $\R$, which can be computed in closed-form for discrete measures with $T$ atoms: 
\begin{align}
\mathcal{W}^2_2(\mu, \nu) &=  \frac{1}{T} \sum_{t=1}^{T} \abs{ \tilde{x}_t - \tilde{y}_t }^2.
\label{intsw}
\end{align}
Here, $\tilde{x}_1\leq...\leq\tilde{x}_T$ and $\tilde{y}_1\leq...\leq\tilde{y}_T$.
In practice, we use Monte Carlo estimation to estimate the integral in \eqref{sw}. Using the slicing method applied to the Wasserstein distance reduces the complexity of computing the OT distance to $O(KT \log T)$ where $K$ is the number of projections and $T$ is the number of samples (atoms per measure) \cite{kolouri2018sliced}.

% The Figure \ref{pr456} presents a simplified example of one projection of a discrete measure from $\mathbb{R}^{2}$ to $\R$.
% \begin{figure}[H]
% 	\centering
%     \includegraphics[width=12cm]{Method/diagram-20210909 (1) (1).png}
%     \caption{Simplified schematic representation of the slicing method. This example contains three state-trajectories (probability distributions) and a unique projection ($\mathcal{P}_1$). }
%     \label{pr456}
% \end{figure}

\subsection{Sliced Multi-Marginal Optimal Transport}
\label{mmsw}

We now describe the sliced multi-marginal distance which allows to compare more than two probability measures \cite{cohen2021sliced} .

 Consider $P$ discrete measures $\mu_1, ..., \mu_P \in M(\mathbb{R})$ with $T$ atoms, the multi-marginal Monge-Wasserstein distance is defined as
\begin{align}
\mathcal{MW}^2_2( \mu_1, ..., \mu_P) &=\frac{1}{N} \sum_{t=1}^N  \lambda_t \lvert \tilde{x}_t^{(p)} - \sum_{j=1}^{P} \lambda_j \tilde{x}_j^{(j)} \rvert ^2, \label{cfmmmwd}
\end{align} where each measure $\mu_1, ..., \mu_P$ can be written as $\frac{1}{N} \sum_{t=1}^N \delta_{\tilde{x}_t^{(p)}} $. The $\tilde{x}_t$s correspond to the sorted values of the atoms, such that $\tilde{x}_1^{(p)} \leq \tilde{x}_2^{(p)} \leq ... \leq \tilde{x}_N^{(p)}, p=1,...P$. $\sum_{p=1}^P \lambda_p x_p$ computes the Euclidean barycenter of the aligned samples and $\lambda \in \Delta_P$ where $\Delta_P$ is a simplex of dimension $P$. 

Using this closed form of the 1D multi-marginal Monge--Wasserstein  \eqref{cfmmmwd}, we can now formulate the sliced multi-marginal Monge--Wasserstein distance that will be useful when dealing with higher-dimensional problems. We define it as follows: 
\begin{align}
\mathcal{SMW}_2^2 (\mu_1, ..., \mu_P) = \frac{1}{Vol(S^{d-1})} \int_{S^{d-1}} \mathcal{MW}^2_2 (\mathcal{P}_{\vtheta \# \mu_1}, ..., \mathcal{P}_{\vtheta \# \mu_P}) d\vtheta.
\label{smmmwhighdims}
\end{align}
In practice, we evaluate \eqref{smmmwhighdims} using Monte Carlo estimation. 

\subsection{Imitation Learning}

We frame continuous control as a Markov decision process $(\gS, \gA, P, R, \gamma, s_0)$ where $\gS$ is the state space the agent lives on, $\gA$ is the action space, $P$ is the transition function such that $\vs_{t+1}\sim P(\vs_t,\va_t)$,  $R:\gS\times \gA \rightarrow \R$ is the reward function, $\gamma$ is the discount factor and $s_0$ is the initial state. 

We consider imitation learning problems where agents need to solve a task given access to a set of expert demonstrations from the expert policy, but without having access to environment rewards. To address this challenge, one option is to infer pseudo-rewards based on the expert demonstrations, and then train agents by reinforcement learning on such pseudo-rewards \cite{Englert2013,DBLP:journals/corr/abs-2006-04678}. 

\section{Methodology}
\label{met}
In this section, we present the main methodology and contribution contained in this work. 

Our primary objective is to investigate and compare two different techniques to dealing with multiple expert demonstrations. The first technique, which is used in recent works \cite{DBLP:journals/corr/abs-2006-04678, zhao2020compressed}, consists in using a concatenation and sampling method and employs the pairwise optimal transport distances (see Section \ref{scotilmeth}), whilst the second technique operates within a multi-marginal setting and relies on multi-marginal optimal transport distances (see Section \ref{smotilmeth}). 

In order to asses the performance of both methods, we implemented two algorithms built on the same imitation learning backbone but with expert demonstrations treated as explained above. We call them \textit{sliced concatenated optimal transport imitation learning} (SCOTIL) and \textit{sliced multi marginal optimal transport imitation learning} (SMMOTIL). These models are based on PWIL algorithm \cite{DBLP:journals/corr/abs-2006-04678}: they use IRL and OT tools to formulate the pseudo-rewards. However, they use a different distance between trajectories and SMMOTIL proposes a different way to combine the experts.

In both approaches, we interpret the expert and agent trajectories as discrete measures, where each measure contains $t$ atoms, one atom for each time step of a trajectory. In imitation learning, the atoms correspond to the states $\vs_t$ denoted $\vs_t^a$ for the agent trajectory and $\vs_{t,p}^e$ for the $p$th expert trajectory. We denote the agent trajectory as $\mu^a = \sum_{i=1}^{t} \delta_{\vs_{t}^a}$ and the expert trajectories as $\mu^e_p = \sum_{i=1}^{t} \delta_{\vs_{t,p}^e}$.

\subsection{Sliced Concatenated Optimal Transport Imitation Learning (SCOTIL)}

\label{scotilmeth}

Sliced concatenated optimal transport imitation learning uses a concatenation and sampling method to combine multiple expert state (-action) trajectories into a single one. Going from a multi-expert setting to a single-(concatenated-)expert setting enables computational savings, as multi-marginal distances are usually challenging to compute. 

The key idea of the concatenation and sampling method is as follows: given $P$ discrete measures $\mu_1^{e}, \mu_2^{e}, ..., \mu_P^{e}$ corresponding to the expert distributions, we concatenate all their states together to obtain a single measure $\mu^{e}$ that contains $tP$ atoms, $t$ being the number of atoms in each measure $\mu_p^{e}$. Then, we use a sub-sampling method to obtain an averaged expert demonstration, i.e., a discrete measure $\mu^{e}$ containing $t$ atoms.

\begin{figure}[H]
    \centering
    \includegraphics[width=12cm]{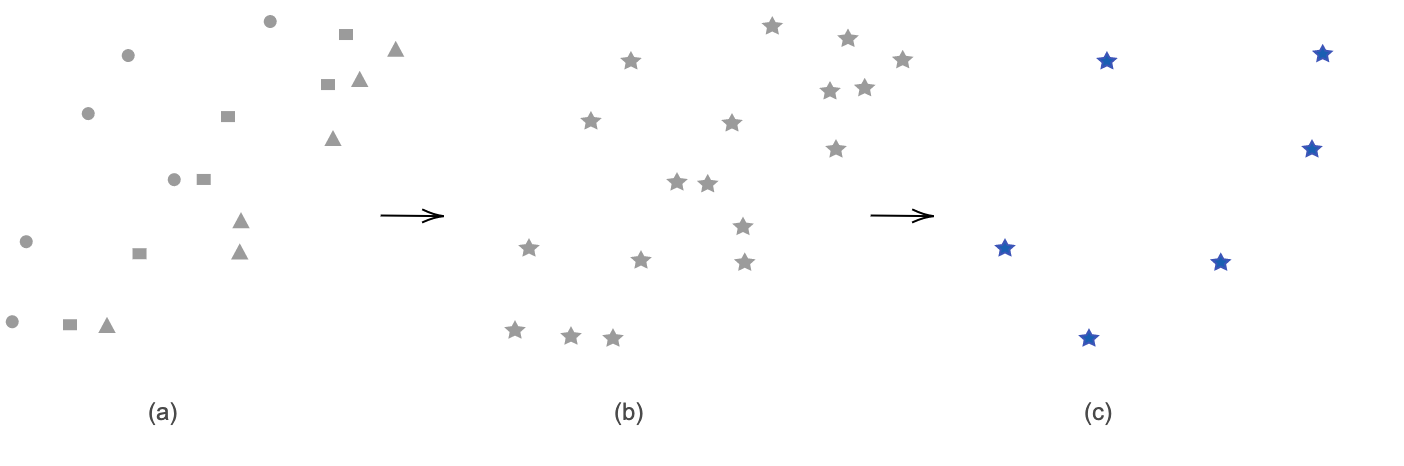}
    \caption{Schematic representation of SCOTIL's concatenation and sampling method. (a) shows three expert demonstrations, (b) shows the concatenated state-trajectory, (c) shows the sampled trajectory that we use as the expert demonstration.}
    \label{pwilschemeexp}
\end{figure}

Once the concatenated expert demonstration is built, we train the agent using a Deep Q-Network (DQN) algorithm (\citet{mnih2013playing}) with pseudo-rewards computed based on the concatenated demonstration and agent rollouts $\mu^{a}$. We leverage the sliced-Wasserstein between the agent rollout and the concatenated expert trajectory to define reward signals. Following \citet{cohen2021sliced}, we define them as
\begin{align}
r(\vs_t^a) = \frac{1}{K}\sum_{k=1}^K |\langle \vs^a_t- \vs^e_{\eta_{k,t}}, \vtheta_k\rangle|,
\label{rew1d}
\end{align}
where $K$ is the number of projections, $\eta_{p,t}$ is the index of the atom in the expert trajectory aligned with $\vs^a_t$ after projecting onto $\vtheta_k$. We note that the sum of rewards equals the squared sliced-Wasserstein distance between agent and expert, so that
\begin{align}
\sum_{t=1}^T r(\vs_t) = \mathcal{SW}^2(\mu^a, \mu^e).
\end{align}
The sliced-Wasserstein distance is estimated with Monte-Carlo. 

\subsection{Sliced Multi-Marginal Optimal Transport Imitation Learning (SMMOTIL)}

\label{smotilmeth}

In the following, we propose an alternative technique to the commonly used method to deal with multiple expert trajectories. Instead of concatenating the demonstrations, we use multi-marginal tools and compare the discrepancy between all the expert state-trajectories $\mu_1^{e}, \mu_2^{e}, ..., \mu_P^{e}$ and the DQN-trained agent rollout $\mu^{a}$. The reward signal can be formulated similarly to \eqref{rew1d}, as derived in \citet{cohen2021sliced},
\begin{align}
r_{t,p}(\vs_t^a, S) &= \frac{1}{PK} \sum_{k=1}^K \Big| \langle \vs_t^{a} - \frac{1}{P} \sum_{j=1}^{P+1} \vs^{(j)}_{\eta_{p,j,k(t)}}, \vtheta_k \rangle\Big|^{2}.
\label{rewhd}
\end{align}
In this model, we note that the sum of rewards is defined by the squared sliced multi-marginal Monge--Wasserstein
\begin{align}
\sum_{t=1}^T  r_{t,p}(\vs_t^a, S)  = \mathcal{SMW}^2(\mu^a, \mu^e_1, ..., \mu^e_P)
\end{align}
between the agent and the experts. The sliced multi-marginal Monge--Wasserstein is estimated using Monte Carlo. 

\begin{figure}[H]
    \centering
    \includegraphics[width=13cm]{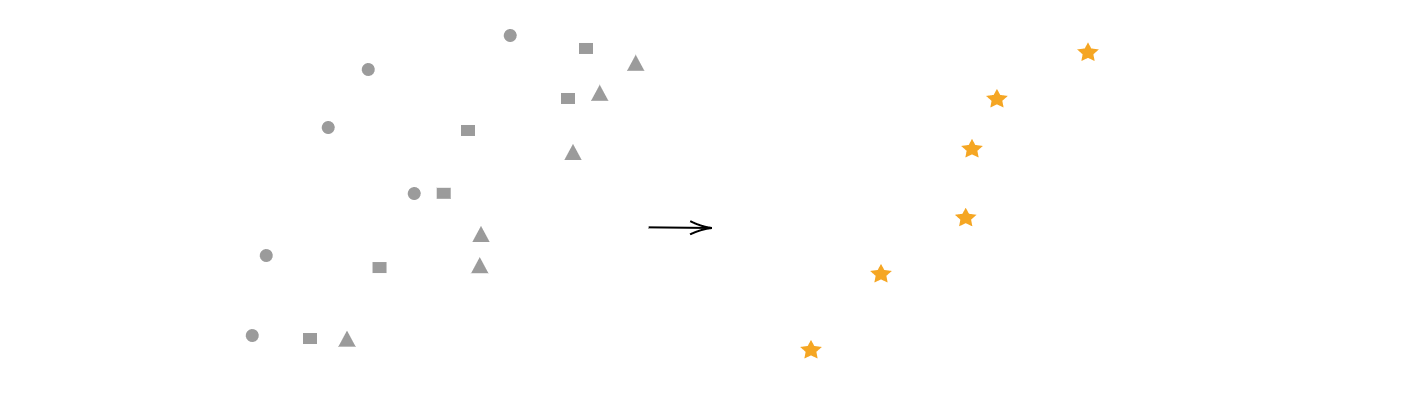}
    \caption{Schematic representation of the barycenter averaging method. It illustrates the geometric averaging of three expert demonstrations}
    \label{smmotexplanatoryplot}
\end{figure}

In theory, minimizing SMMOTIL's loss function is equivalent to minimizing the sliced-Wasserstein between the agent's rollout and the sliced-Wasserstein barycenter of the expert trajectory. The barycenter averaging is based on a geometric averaging method, which is expected to lead to a smoother expert trajectory than the concatenation and sampling method as illustrated in Figure \ref{smmotexplanatoryplot}; more details are given in \cite{BRPP15,  bigot2017characterization,cohen2021sliced}. 

\subsection{Algorithmic Details}

SMMOTIL and SCOTIL present differences in their way to treat the expert demonstrations. However, they have the same core imitation learning approach. In both implemented models, we use IRL: the agent is a DQN  (\cite{mnih2013playing}) that learns from the  imitation rewards defined in  \eqref{rew1d} and \eqref{rewhd}, respectively. The rewards are computed using the sliced distances (in pairwise-marginal or multi-marginal settings). Details about parameters are given in Appendix \ref{impldet}. 

%%%%%%%%%%%%%%%
\section{Experimental Results}

With the aim of evaluating and comparing different averaging methods with episodic rewards  as metric (as explained in Section \ref{met}), we conduct a set of experiments on Open AI gym control tasks. We gather diverse optimal expert demonstrations: five with a different length and five others with a different mass using a DQN\footnote{The length refers to the length of the pendulum and pole and the mass refers to the mass of the pendulum and cart for Pendulum-v0 and CartPole-v0 respectively.}. Then, for each environment, we carry out two experiments: in the first one, an agent learns from the five diverse-length-experts and in the second one, an agent learns from the five diverse-mass-experts. For each experiment and each environment, we compare the efficiency of SMMOTIL and SCOTIL. A summary of the conducted experiments is given in Table \ref{level3exp}. 

\begin{figure}[H]
  \begin{center}
    \begin{minipage}[b]{0.4\textwidth}
    \includegraphics[width=\textwidth]{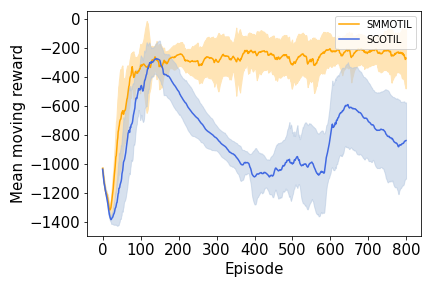}
    \caption*{Diverse lengths in the Pendulum-v0}
    \vspace{0.5cm}
  \end{minipage}
      \begin{minipage}[b]{0.4\textwidth}
    \includegraphics[width=\textwidth]{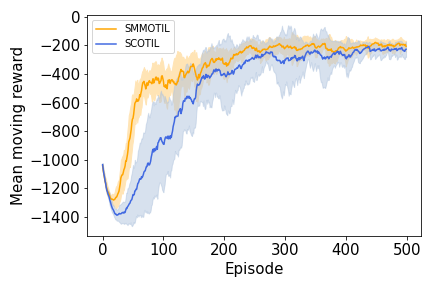}
    \caption*{Diverse masses in Pendulum-v0}
    \vspace{0.5cm}
  \end{minipage}
    \begin{minipage}[b]{0.4\textwidth}
    \includegraphics[width=\textwidth]{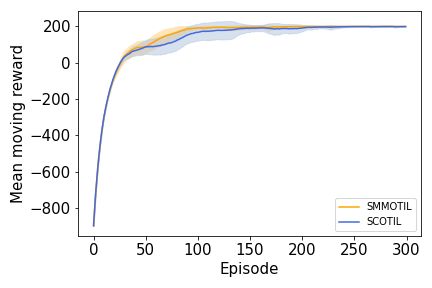}
    \caption*{Diverse lengths in CartPole-v0}
  \end{minipage}
     \begin{minipage}[b]{0.4\textwidth}
    \includegraphics[width=\textwidth]{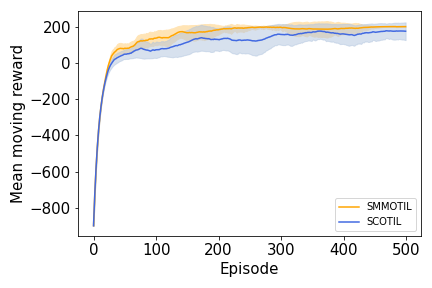}
    \caption*{Diverse masses in CartPole-v0 }
  \end{minipage}
  \end{center}
  \caption{Experimental results}
  \label{l3}
\end{figure}

\begin{table}[H]
\begin{center}
\begin{tabular}{p{.3\textwidth}p{.3\textwidth}p{.3\textwidth}}
\toprule
\rowcolor{black!20} & Pendulum-v0 & CartPole-v0 \\
\hline
Agent length & 1 & 0.5 \\
Experts lengths & 0.3, 0.5, 1.2, 1.5, 1.7 & 0.1, 0.3, 1.2, 1.5, 2.0 \\
Agent mass & 1 & 1  \\
Experts masses & 0.1, 0.6, 1.2, 1.8, 2.0 & 0.001, 0.5, 2.1, 5.0, 8.0 \\
\hline
\end{tabular}
\captionof{table}{Summary of the different experts used}
\label{level3exp}
\end{center}
\end{table} 
We display the experimental results in Figure \ref{l3}: each graph summarizes the results from 10 experiments; the solid line corresponds to the mean moving reward per episode and the shaded area to the corresponding standard deviation. 

We observe that, in all four experimental result plots, SMMOTIL's mean rewards are constantly higher than SCOTIL's mean rewards through the episodes. Also, SCOTIL's method presents a higher variance in rewards, and is unstable.

When the expert trajectories are diverse, the concatenation and sampling method (used in SCOTIL) leads to a noisy expert demonstration, which we expect to explain the fact that SCOTIL is under-performing.  

\section{Conclusion} 

In this paper, we compared two different techniques to combining expert demonstrations in imitation learning, both leveraging optimal transport tools. The first model combines them by concatenation and sampling and uses pairwise-marginal optimal transport tools; the second model uses multi-marginal distances to compute the discrepancy between the state-trajectories of all the experts and the agent's rollout without concatenating the demonstrations, which is a proxy for comparing the agent's rollout to the sliced Wasserstein barycenter of the experts \cite{cohen2021sliced}.
The results obtained on OpenAI Gym control demonstrates the suitability of using a multi-marginal approach for combining experts, especially when the experts used to teach to the agent are diverse (e.g., due to variations in environments such as pendulum's and carpole's lengths and masses). 
To conclude, this work introduces a new method that enables using multiple expert demonstrations in imitation learning via sliced multi-marginal optimal transport.

It will be interesting to further verify our finding on higher dimensional environments such as MuJuCo control tasks in future work. It will be also interesting to consider other metrics like (sliced-)Gromov-Wasserstein distances which would allow to compare agents and experts living on different spaces, similarly to \cite{cohen2021aligning}, but with an IRL approach instead.

\subsection*{Acknowledgments}
SC was supported by the Engineering and Physical Sciences Research Council (grant number EP/S021566/1).

\bibliographystyle{apalike}
\bibliography{references}

\begin{thebibliography}{}

\bibitem[Bighashdel et~al., 2021]{bighashdel2021deep}
Bighashdel, A., Meletis, P., Jancura, P., and Dubbelman, G. (2021).
\newblock Deep adaptive multi-intention inverse reinforcement learning.
\newblock In {\em ECML PKDD 2021}.

\bibitem[Bigot and Klein, 2017]{bigot2017characterization}
Bigot, J. and Klein, T. (2017).
\newblock Characterization of barycenters in the wasserstein space by averaging
  optimal transport maps.

\bibitem[Bonneel et~al., 2015]{BRPP15}
Bonneel, N., Rabin, J., Peyr{\'e}, G., and Pfister, H. (2015).
\newblock {Sliced and Radon Wasserstein Barycenters of Measures}.
\newblock {\em Journal of Mathematical Imaging and Vision}, 51(1):22–45.

\bibitem[Cohen et~al., 2021a]{cohen2021sliced}
Cohen, S., Kumar, K. S.~S., and Deisenroth, M.~P. (2021a).
\newblock Sliced multi-marginal optimal transport.
\newblock {\em arXiv:2102.07115}.

\bibitem[Cohen et~al., 2021b]{cohen2021aligning}
Cohen, S., Luise, G., Terenin, A., Amos, B., and Deisenroth, M.~P. (2021b).
\newblock Aligning time series on incomparable spaces.
\newblock {\em arXiv:2006.12648}.

\bibitem[Dadashi et~al., 2020]{DBLP:journals/corr/abs-2006-04678}
Dadashi, R., Hussenot, L., Geist, M., and Pietquin, O. (2020).
\newblock Primal wasserstein imitation learning.
\newblock {\em CoRR}, abs/2006.04678.

\bibitem[Englert et~al., 2013]{Englert2013}
Englert, P., Paraschos, A., Peters, J., and Deisenroth, M.~P. (2013).
\newblock Model-based imitation learning by probabilistic trajectory matching.
\newblock In {\em Proceedings of the IEEE International Conference on Robotics
  and Automation}.

\bibitem[Florence et~al., 2021]{florence2021implicit}
Florence, P., Lynch, C., Zeng, A., Ramirez, O., Wahid, A., Downs, L., Wong, A.,
  Lee, J., Mordatch, I., and Tompson, J. (2021).
\newblock Implicit behavioral cloning.
\newblock {\em arXiv:2109.00137}.

\bibitem[Goecks et~al., 2020]{goecks2020integrating}
Goecks, V.~G., Gremillion, G.~M., Lawhern, V.~J., Valasek, J., and Waytowich,
  N.~R. (2020).
\newblock Integrating behavior cloning and reinforcement learning for improved
  performance in dense and sparse reward environments.
\newblock {\em arXiv:1910.04281}.

\bibitem[Jarboui and Perchet, 2021a]{jarboui2021generalised}
Jarboui, F. and Perchet, V. (2021a).
\newblock A generalised inverse reinforcement learning framework.
\newblock {\em arXiv:2105.11812}.

\bibitem[Jarboui and Perchet, 2021b]{jarboui2021offline}
Jarboui, F. and Perchet, V. (2021b).
\newblock Offline inverse reinforcement learning.
\newblock {\em arXiv:2106.05068}.

\bibitem[Karnan et~al., 2021]{karnan2021voila}
Karnan, H., Warnell, G., Xiao, X., and Stone, P. (2021).
\newblock Voila: Visual-observation-only imitation learning for autonomous
  navigation.
\newblock {\em arXiv:2105.09371}.

\bibitem[Kim et~al., 2021]{kim2021transformerbased}
Kim, H., Ohmura, Y., and Kuniyoshi, Y. (2021).
\newblock Transformer-based deep imitation learning for dual-arm robot
  manipulation.
\newblock {\em arXiv:2108.00385}.

\bibitem[Kolouri et~al., 2019]{kolouri2018sliced}
Kolouri, S., Pope, P.~E., Martin, C.~E., and Rohde, G.~K. (2019).
\newblock Sliced wasserstein auto-encoders.
\newblock In {\em International Conference on Learning Representations}.

\bibitem[Mnih et~al., 2013]{mnih2013playing}
Mnih, V., Kavukcuoglu, K., Silver, D., Graves, A., Antonoglou, I., Wierstra,
  D., and Riedmiller, M. (2013).
\newblock Playing atari with deep reinforcement learning.

\bibitem[Ng and Russell, 2000]{Ng00algorithmsfor}
Ng, A.~Y. and Russell, S. (2000).
\newblock Algorithms for inverse reinforcement learning.
\newblock In {\em in Proc. 17th International Conf. on Machine Learning}, pages
  663--670. Morgan Kaufmann.

\bibitem[Papagiannis and Li, 2020]{papagiannis2020imitation}
Papagiannis, G. and Li, Y. (2020).
\newblock Imitation learning with sinkhorn distances.
\newblock {\em arXiv:1906.08113}.

\bibitem[Pomerleau, 1991]{6796843}
Pomerleau, D.~A. (1991).
\newblock Efficient training of artificial neural networks for autonomous
  navigation.
\newblock {\em Neural Computation}, 3(1):88--97.

\bibitem[Ratliff et~al., 2006]{10.5555/2976456.2976601}
Ratliff, N., Bradley, D., Bagnell, J.~A., and Chestnutt, J. (2006).
\newblock Boosting structured prediction for imitation learning.
\newblock In {\em Proceedings of the 19th International Conference on Neural
  Information Processing Systems}, NIPS'06, page 1153–1160, Cambridge, MA,
  USA. MIT Press.

\bibitem[Russell, 1998]{10.1145/279943.279964}
Russell, S. (1998).
\newblock Learning agents for uncertain environments (extended abstract).
\newblock In {\em Proceedings of the Eleventh Annual Conference on
  Computational Learning Theory}, COLT' 98, page 101–103, New York, NY, USA.
  Association for Computing Machinery.

\bibitem[Stepputtis et~al., 2019]{stepputtis2019imitation}
Stepputtis, S., Campbell, J., Phielipp, M., Baral, C., and Amor, H.~B. (2019).
\newblock Imitation learning of robot policies by combining language, vision
  and demonstration.
\newblock {\em arXiv:1911.11744}.

\bibitem[Wu et~al., 2021]{wu2021textgail}
Wu, Q., Li, L., and Yu, Z. (2021).
\newblock Textgail: Generative adversarial imitation learning for text
  generation.
\newblock {\em arXiv:2004.13796}.

\bibitem[Xiao et~al., 2019]{xiao2019wasserstein}
Xiao, H., Herman, M., Wagner, J., Ziesche, S., Etesami, J., and Linh, T.~H.
  (2019).
\newblock Wasserstein adversarial imitation learning.
\newblock {\em arXiv:1906.08113}.

\bibitem[Zhao and Lou, 2020]{zhao2020compressed}
Zhao, N. and Lou, B. (2020).
\newblock Compressed imitation learning.
\newblock {\em arXiv:2009.11697}.

\bibitem[Zheng et~al., 2019]{zheng2019simultaneous}
Zheng, B., Zheng, R., Ma, M., and Huang, L. (2019).
\newblock Simultaneous translation with flexible policy via restricted
  imitation learning.
\newblock {\em ACL Anthology}.

\end{thebibliography}

\appendix
\section{Appendix}

\subsection{Implementation details}
\label{impldet}
\begin{center}
\begin{tabular}{p{.5\textwidth}p{.5\textwidth}}
\toprule
\rowcolor{black!20} Parameter & Value  \\
\hline
Maximum steps per episode & 200 \\
Number of expert demonstrations & 5 \\
Number of projections for the slicing & 50 \\
Learning rate & $1e^{-3}$\\
Optimizer & Adam \\
Discount factor & 0.99 \\
Replay memory size & 2000\\
Batch size & 32 \\
Initial $\varepsilon$'s value for the $\varepsilon$-greedy policy & 1 \\
Final $\varepsilon$'s value for the $\varepsilon$-greedy policy & 0.01 \\
\hline
\end{tabular}
\captionof{table}{Implementation details}
\label{idil}
\end{center}

\end{document}